\documentclass{article} %
\usepackage{iclr2023_conference_modified,times}

\usepackage{amsmath,amsfonts,bm}

\def\eqref#1{equation~\ref{#1}}

\def\1{\bm{1}}

\DeclareMathAlphabet{\mathsfit}{\encodingdefault}{\sfdefault}{m}{sl}
\SetMathAlphabet{\mathsfit}{bold}{\encodingdefault}{\sfdefault}{bx}{n}

\usepackage[T1]{fontenc}
\usepackage{hyperref}
\usepackage{url}
\usepackage{graphicx}
\usepackage{tikz}
\usepackage{comment}
\usepackage{amsmath,amssymb} %
\usepackage{color}
\usepackage{hyperref}
\usepackage{url}
\usepackage[T1]{fontenc}    %
\usepackage{booktabs}       %
\usepackage{amsfonts}       %
\usepackage{nicefrac}       %
\usepackage{microtype}      %
\usepackage{color}
\usepackage{amsmath}
\usepackage{tikz}
\usetikzlibrary{arrows, positioning, automata}
\usepackage{amssymb}
\usepackage{pifont}
\usepackage{graphicx}
\usepackage{subcaption}
\usepackage{multirow}
\usepackage[normalem]{ulem}
\usepackage{enumitem}
\usepackage{wrapfig}
\usepackage{listings}

\definecolor{codegreen}{rgb}{0.36, 0.54, 0.66}
\definecolor{codegray}{rgb}{1,1,1}
\definecolor{codepurple}{rgb}{0.58,0,0.82}
\definecolor{bluepigment}{rgb}{0.2, 0.2, 0.6}
\definecolor{amethyst}{rgb}{0.6, 0.4, 0.8}
\definecolor{backcolour}{rgb}{1,1,1}
\lstdefinestyle{mystyle}{
    backgroundcolor=\color{backcolour},   
    numberstyle=\tiny\color{codegray},
    basicstyle=\ttfamily\tiny,
    breakatwhitespace=false,         
    breaklines=true,                 
    captionpos=b,                    
    keepspaces=true,                 
    numbers=left,                    
    numbersep=5pt,                  
    showspaces=false,                
    showstringspaces=false,
    showtabs=false,                  
    tabsize=2
}
\usepackage{caption}
\usepackage[scr=boondoxo]{mathalfa}
\usepackage[scaled=.8]{beramono}
\usepackage{color, colortbl,booktabs}
\usepackage{algorithm}
\usepackage[noend]{algpseudocode}
\algnewcommand{\LineComment}[1]{\State \(\triangleright\) #1}

\usepackage{caption}

\title{Projected Subnetworks Scale Adaptation}

\iclrfinalcopy

\author{Siddhartha Datta
\\
University of Oxford\\
\And
Nigel Shadbolt \\
University of Oxford \\
}

\newcommand{\acronym}{SNP}

\usepackage{cases}
\usepackage{amsmath,amsthm}

\begin{document}
\maketitle

\begin{abstract}
Large models support great zero-shot and few-shot capabilities.
However, updating these models on new tasks can break performance on previous seen tasks and their zero/few-shot unseen tasks.
Our work explores how to update zero/few-shot learners such that they can maintain performance on seen/unseen tasks of previous tasks as well as new tasks.
By manipulating the parameter updates of a gradient-based meta learner as the projected task-specific subnetworks, 
we show improvements for large models to retain seen and zero/few shot task performance in online settings.
\end{abstract}

\section{Introduction}
\label{1}

The adaptation of deep neural networks have practical importance.
It enables models to adapt to varying test-time distributions, attributed to shifts in time, person, environment, etc.
The more difficult adaptation cases arise when there may be no clear task boundaries, when the task was not seen during training, and only few/zero samples are available to update a model. 
To tackle adaptation broadly,
given a \textit{base learner}
optimizing its inner objective with respect to its assigned task,
a \textit{meta learner} 
computes the update to the base learner such that it optimizes its outer objective across a distribution of tasks \citep{hospedales2021meta}.
Scaling the size of models and training data have recently demonstrated comparable zero/few-shot capabilities (e.g. GPT-3 \citep{https://doi.org/10.48550/arxiv.2005.14165}, CLIP \citep{radford2021learning}, Chinchilla \citep{hoffmann2022training}).

Retaining this zero/few-shot capability becomes a challenge in an online setting.  
Prior continual learning methods \citep{Lange2019ContinualLA} aim to retain performance on both prior and subsequent tasks, but do not evaluate the retention of zero/few-shot task performance. 
\citet{ilharcopatching} proposed an online algorithm that fine-tunes a large vision-language model on a new task, and performs well on the previous zero/few-shot tasks and the seen fine-tuned task.

Task-specific representations within a large model can be difficult to disentangle and manipulate. 
Identifying and freezing subnetworks (e.g. APD \citep{Yoon2020Scalable}, WSN \citep{pmlr-v162-kang22b}) can help mitigate forgetting. 
A meta learner projects its representations onto a base parameter space.
For a gradient-based meta learner, the meta learner and base parameters reside in the same parameter space. 
By optimizing meta parameters alike to gradient-based meta learning,
we can project the task-specific representations (subnetworks) in the meta parameters to interpolatable, equidimensional base parameters (subnetworks) in one parameter space. 
Our proposed method, Subnetwork Projection (\acronym), trains a meta learner to maximize the distance that the meta parameters can drift while returning the same base parameters.
\acronym++ also stores a memory buffer to access and manipulate the base parameters.
In addition to subnetwork addition/removal,
this paradigm enables 
zero/few-shot subnetworks,
interpolatable subnetworks, 
and other network editing possibilities. 

\noindent\textbf{Contributions. }
Subnetwork Projection (\acronym) is the first continual learner designed to retain seen and unseen zero/few-shot performance on both prior and subsequent tasks, outperforming existing baselines.
By projecting subnetworks as equidimensional base parameters in the same space, 
SNP trains a model to sustain greater parameter drift while still retaining access to the original base parameters.
With task-specific samples, SNP++ can manipulate subnetworks encoded in the meta parameters, including adding, removing, combining, or switching subnetworks.

\section{Related Work}

Modifiying neural networks can rely on varying levels of discretization of underlying representations.
This results in different adaptation techniques of models to tasks.

\textbf{Discrete representations. }
Identifying and freezing task-specific subnetworks can minimize forgetting on prior tasks \citep{Yoon2020Scalable, pmlr-v162-kang22b}. 
A concern with this discrete representation is its mutability. 
Once a subnetwork is identified and frozen, it cannot be transformed or switched to a different subnetwork if its source data is unavailable. This would be needed if most of the network is frozen and no leftover nodes are available for a new task. 
Task order affects the subnetwork arrangement, and changes in unfrozen subnetworks values may render the frozen subnetwork inaccurate. 
Network capacity would also be fixed; once all the nodes are frozen, a subnetwork would need to be removed in order for a new task to be learnt. 
Interpolation between subnetworks may not viable due to different shapes. 
Subnetworks could also be constructed as stitchable layers of regular shapes (e.g. linear layer), such as in model stitching \citep{https://doi.org/10.48550/arxiv.2110.14633, bansal2021revisiting} or feature adapters \citep{gao2021clip, chen2022adaptformer}.
These layers would need to be compatible and conditioned on the previous layers. 
Networks can also be modularly generated from architectural components \citep{Andreas_2016_CVPR, https://doi.org/10.48550/arxiv.1601.01705}.

\textbf{Continuous representations. }
Instead of manipulating discrete, composable modules in neural networks, the weights/parameters of the network can be modified.
Combining representations is a common technique, aiding in leveraging transferable properties between tasks as well as conserving capacity.
Regularization-based continual learning strategies
\citep{EWC, 10.5555/3305890.3306093, https://doi.org/10.48550/arxiv.2209.14996}
use regularization terms to update parameters towards the new task while retaining pertinent representations of the prior task. 
Model merging and averaging has also been used to improve generalizability, robustness, and adaptatability in online learning settings \citep{ https://doi.org/10.48550/arxiv.1910.05653,
https://doi.org/10.48550/arxiv.2111.09832,
https://doi.org/10.48550/arxiv.2203.05482,
ilharcopatching}. 
Other than interpolatability,
residing in a continuous space enables these representations to be dynamically-generated. 
Meta learners \citep{pmlr-v70-finn17a, ha2016hypernetworks} query a few samples from the task to compute the updated base parameters.
Large models can support similar few-shot capabilities with parameter-efficient fine-tuning, such as prompt tuning \citep{sanh2022multitask, wei2022finetuned, wei2022emergent, Zhou_2022}.

\section{Grounding subnetwork projection in meta learners}

\begin{table*}[t]
\caption{
Evaluating models fine-tuned on each dataset against each other.
}
\begin{subfigure}{\textwidth}
\centering
\caption{
Measuring the cosine distance between flattened parameters fine-tuned on each dataset against each other. 
}
\includegraphics[width=\textwidth]{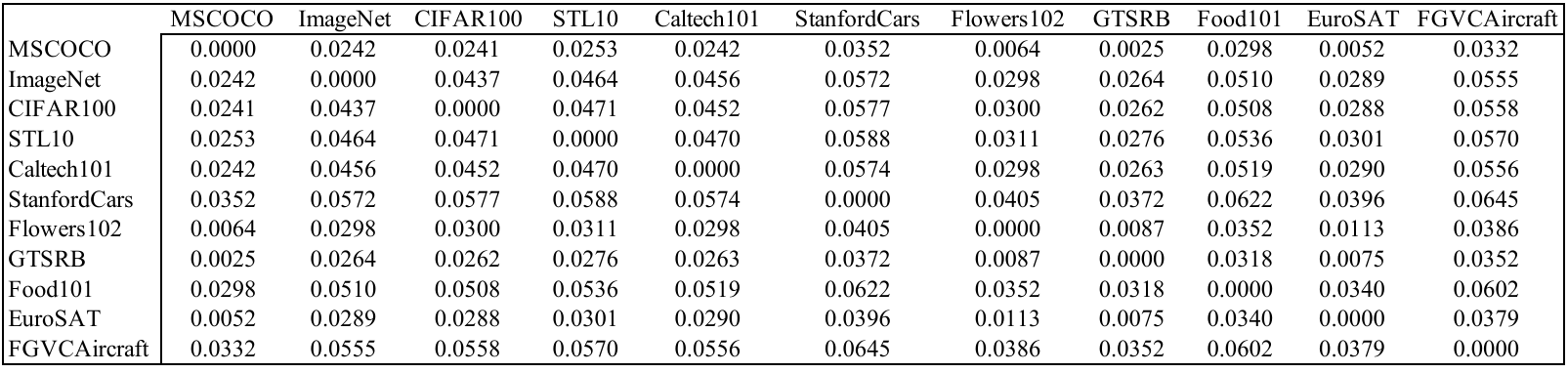}
\label{fig:taba}
\end{subfigure}
\begin{subfigure}{\textwidth}
\centering
\caption{
Measuring the (Zero-shot Top-5 / Few-shot Top-1) accuracy between parameters fine-tuned on each dataset against each other. 
We baseline against the pretrained initialization trained on WIT, 
the finetuned model on MSCOCO (which was then used as the starting point for each subsequently-finetuned model). 
}
\includegraphics[width=\textwidth]{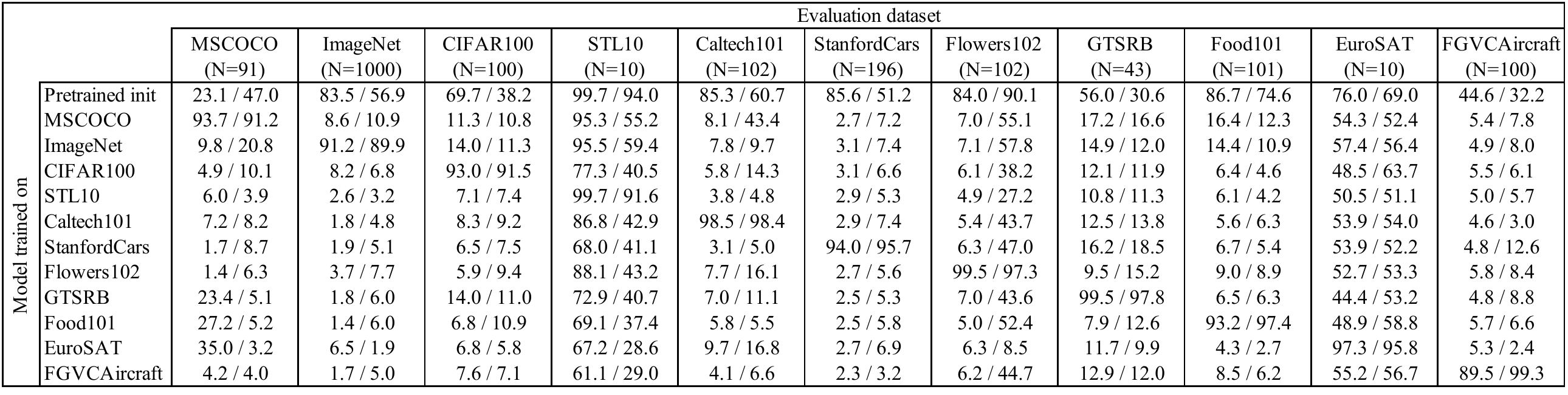}
\label{fig:tabb}
\end{subfigure}
\end{table*}

\begin{table*}
\centering
\caption{
Empirical equivalence between CLIP and MAML(CLIP).
Both networks have the same number of parameters and architecture, but vary by training optimization procedure. They can both be trained to attain comparable (Zero-shot Top-5 / Few-shot Top-1) accuracy.
}
\includegraphics[width=\linewidth]{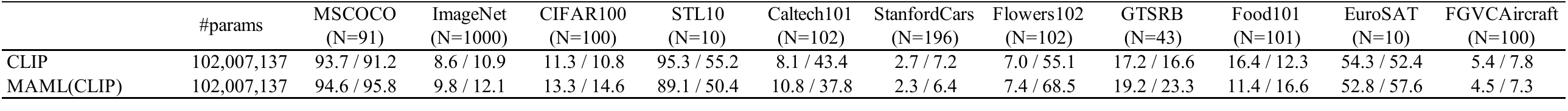}
\label{fig:maml_clip}
\end{table*}

\noindent\textbf{Problem Setup.}
From a task set $\mathbb{T}=\{\mathcal{T}_t\}^{t \in T}$,
a base learner receives $T$ tasks sequentially.
$\mathcal{T}_t = \{ x_{t}, y_{t} \}$ denotes the dataset of the $t$-th task. 
In the online/continual learning setting, 
given loss function $\mathcal{L}$, 
a base learner $\mathscr{f}(\theta_{\textnormal{base}}; x)$ optimizes its parameters $\theta_{\textnormal{base}}$ such that
it can perform well on the $t$-th task while minimizing performance drop on the previous $(t-1)$ tasks.
We further add the requirement of retaining the zero/few-shot performance of unseen tasks $\mathbb{V}_t = \{\mathcal{V}_{t, v}\}^{v \in V}$ of the $t$-th seen task.
Hence, the objective is:
$\theta_{\textnormal{base}}^{*} := \arg\min_{\theta_{\textnormal{base}}} \sum_{t=1}^{T}[\mathcal{L}(\mathscr{f}(\theta_{\textnormal{base}}; x_{t}),  y_{t}) + \sum_{v=1}^{V} \mathcal{L}(\mathscr{f}(\theta_{\textnormal{base}}; x_{v}),  y_{v})] $.
We measure drift between two parameters with the distance function $\texttt{dist}(\theta_0, \theta_1)$.
The experimental setup is described in Section \ref{exp}.

\begin{algorithm}[ht]
\small
\caption{{\bf \texttt{base\textunderscore params}}}
\begin{algorithmic}[1]
\Procedure{\texttt{base\textunderscore params}}{$\theta$, 
$\mathbb{T} = \{ \mathcal{T}_t \}^{t \in T}$,
$K$,
$lr_{\textnormal{base}}$}
\For {$\mathcal{T}_t$ in $\mathbb{T}$}
    \For {$X^K_t$, $Y^K_t$ in $\mathcal{T}_t$}
        \State $\theta_{\textnormal{base}, t} = \theta - lr_{\textnormal{base}} \frac{\partial \mathcal{L}(\theta; X^K_t, Y^K_t)}{\partial \theta}$
    \EndFor
\EndFor
\State \textbf{return} $\{ \theta_{\textnormal{base}, t} \}^{t \in T}$
\EndProcedure
\end{algorithmic}
\label{alg:joint_inference}
\end{algorithm}

\subsection{Task groupings}

The pre-trained model was trained on WebImageText (WIT) \citep{radford2021learning}, a dataset specially gathered with 400M (image, text) pairs, in contrast to 100K in MSCOCO \citep{https://doi.org/10.48550/arxiv.1405.0312}.
From the pre-trained initialization,
we train on the first task of MSCOCO for 50 epochs, until the accuracy is high on MSCOCO and low for all the unseen datasets that the pre-trained model performed well on. 
We reuse 10 datasets used in CLIP's evaluation 
including 
ImageNet \citep{deng2009imagenet},
CIFAR100 \citep{krizhevsky2009learning}, 
STL-10 \citep{coates2011stl10}, 
Caltech-101 \citep{FeiFei2004LearningGV},
Stanford Cars \citep{KrauseStarkDengFei-Fei_3DRR2013},
Oxford Flowers 102 \citep{Nilsback08}, 
GTSRB \citep{Stallkamp2012}, 
Food-101 \citep{bossard14},
EuroSAT \citep{helber2017eurosat},
and FGVC Aircraft \citep{maji13fine-grained}.
While MSCOCO contains (image, caption) pairs, the other datasets contain (image, label) pairs. 
Hence, with the prompt templates
provided by \citet{radford2021learning} for each dataset (e.g. “a photo of a \texttt{{CLASS}}”, we can convert labels to captions.

While the pre-trained initialization performs well across the 11 datasets, the MSCOCO-finetuned model loses many transferable representations from WIT, such that the average zero/few-shot accuracy is low.
MSCOCO is on a much smaller scale than WIT, 
in terms of number of images (100K vs 400M), 
range of classes (e.g. ImageNet labels such as stingray, tarantula, mousetrap are not found in MSCOCO),
and less diverse images (e.g. MSCOCO contains natural and day-to-day scenes, while WIT contains natural scenes, sketches, blurry images, low-res images, texts, websites).
The lifelong goal is to 
gradually increase the average zero/few-shot accuracy across all tasks. 

Given the range of datasets, we evaluate zero/few-shot transferability between them, 
such that learning one dataset will yield high zero/few-shot performance on another dataset. 
First, we fine-tuned CLIP on each dataset from a MSCOCO-finetuned initialization.

In Table \ref{fig:taba}, we measured the cosine distance between each pair of models fine-tuned on two different datasets. 
This infers the spatial distance in the parameter space, which parameters are closer to each other, and which parameters require further gradient updates from the initialization. 
We are able to identify 3 sets of datasets, grouped by distance: (i) $\leq 0.1$, (ii) $0.1-0.3$, (iii) $\geq 0.3$. 
In Table \ref{fig:tabb}, we evaluate the functional distance of each fine-tuned model
by computing the zero/few-shot performance per model on each dataset.

Based on the relational analysis between seen and unseen tasks, given the distance between models fine-tuned on each dataset (relative cosine distance in the parameter space), and given the zero/few-shot performance of each fine-tuned model, some clear task groupings can be identified. The order for the seen (unseen) tasks is: MSCOCO (FGVCAircraft, EuroSAT, Food101) → ImageNet (STL10, StanfordCars) → Caltech101 (CIFAR100, GTSRB, Flowers102).

\begin{table}
\centering
\caption{
Evaluating distance measurements between meta parameters and base parameters.
}
\begin{subfigure}{\textwidth}
\centering
\includegraphics[width=\linewidth]{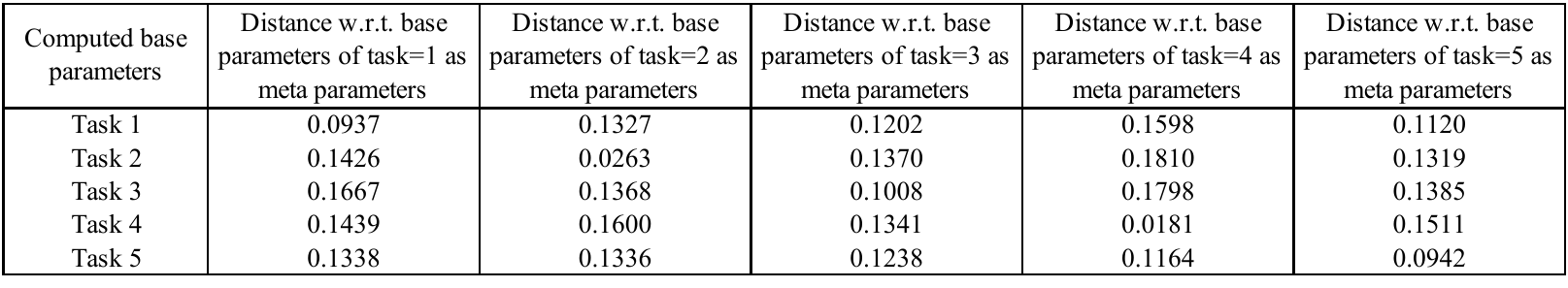}
\caption{
After drifting the meta parameters to each task's base parameters, 
the newly-computed base parameters for each task has minimal drift with respect to the original base parameters.
}
\label{fig:dist3}
\end{subfigure}
\begin{subfigure}{.3\textwidth}
\centering
\includegraphics[width=\textwidth]{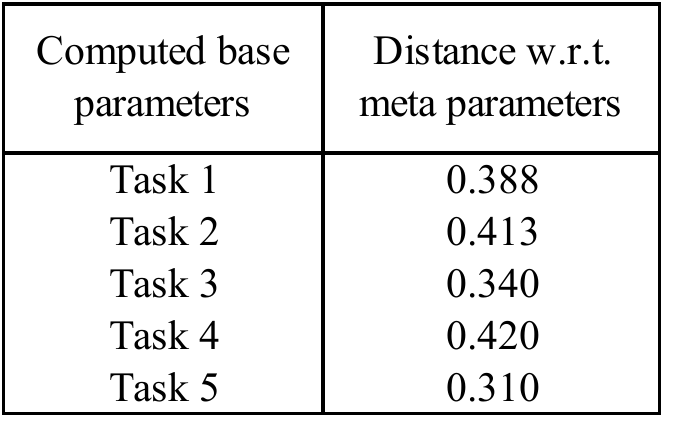}
\caption{
Measuring the Euclidean distance between the original meta parameters of MAML(CLIP) and each task's base parameters is indicative of the subspace radius.
}
\label{fig:dist1}
\end{subfigure}
\hspace{0.1cm}
\begin{subfigure}{.5\textwidth}
\centering
\includegraphics[width=\textwidth]{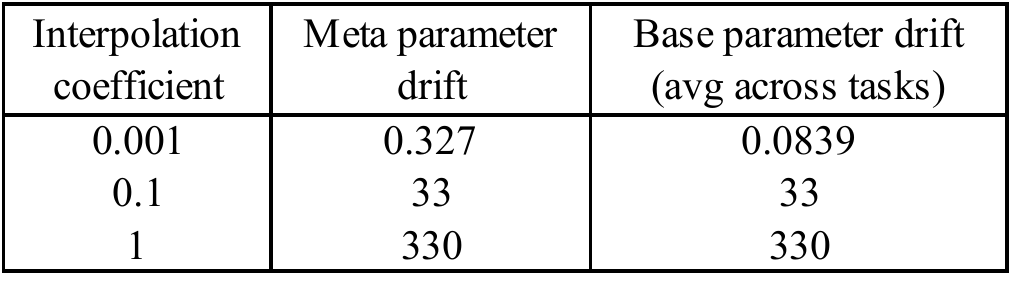}
\caption{
Interpolating the meta parameters against randomly-sampled distant parameters,
we find that the closer the interpolated meta parameters, 
the closer the Euclidean distance, 
and within certain limits of drift, 
the base parameters would drift minimally. 
}
\label{fig:dist2}
\end{subfigure}
\end{table}

\subsection{Disentangling model representations as base parameters}

Meta learners are trained to develop zero/few-shot capabilities. 
Given meta parameters, 
some meta learners map input distributions to base parameters. 
The base parameters are task-specific representations, and dynamically generated with respect to the task. 
In the case of gradient-based meta learners, the meta parameters and base parameters reside in the same parameter space.
A gradient is computed with respect to the new task's input distribution and applied to the meta parameters, and this returns the base parameters. 
As a result, we can project the task-specific representations and subnetworks within the meta parameters to the base parameter space,
and use a gradient-based meta learner to retain the same model architecture and output space as training a model without a meta learning training procedure. 
We train CLIP with MAML's \citep{finn2017modelagnostic} training procedure for 10,000 epochs, meta learning rate and base learning rate of 0.0005. 
To retain the same scale of model and data, we use the same CLIP architecture and capacity, 
and retain the same dataset size by training MAML(CLIP) on Split-MSCOCO \citep{10.5555/3495724.3497128} (which organizes labelled MSCOCO into 5 tasks: transport, animals, sports, food, interior). 
We train MAML(CLIP) until it attains similar zero/few-shot performance as CLIP (Table \ref{fig:maml_clip}).

\subsection{Drift in a meta parameter's subspace}

When updating the meta learning parameters $\theta$ on a new task, $\theta$ may drift by a certain distance to $\theta^{'}$.
Given that the base parameters $\{\theta_{\textnormal{base}, t}\}^{t \in T}$ and $\theta$ reside in the same parameter space, 
we evaluate how far  $\theta$ can drift while being able to compute the same (or within bounded error $\varepsilon$) base parameters $\texttt{dist}(\theta_{\textnormal{base}, t}, \theta^{'}_{\textnormal{base}, t}) \leq \varepsilon$.

From Table \ref{fig:dist1}, we first measure the Euclidean distance between the original meta parameters and their MAML-computed base parameters. This informs the approximate radius of the parameter subspace. 
In Table \ref{fig:dist3}, we test whether drifting the meta parameter to each task's base parameter $\theta = \theta_{\textnormal{base}, t}$ will still be able to compute the same task base parameters. 
Relative to the subspace radius,
we find that the base parameters can indeed be re-located if the meta parameter is drifted to the end-points of the subspace. 
Given that the base parameters can be located if the drift is within the bounds of the subspace, we next evaluate whether the base parameters can be located if the drift exceeds the bounds of the subspace. 
In Table \ref{fig:dist2}, we evaluate $S=1000$ random parameters, and interpolate between this random parameter $\theta_{\textnormal{rand}, s}$ and the meta parameter $\theta$ to return an interpolated meta parameter $\theta_{\textnormal{int}} = (1-r)\theta + r\theta_{\textnormal{rand}, s}$ with interpolation coefficient $r$.
We find that for varying interpolation coefficients (and thus varying Euclidean distances), once the Euclidean distance increases substantially, then the computed base parameters drift in a similar fashion from the original base parameters. 
As a result, we are interested in maximizing the radius of the parameter subspace in which the meta parameter can drift, while returning the same base parameters (within bounded error).

\begin{algorithm}[ht]
\small
\caption{{\bf \texttt{adaptive\textunderscore beta}}}
\begin{algorithmic}[1]
\Procedure{\texttt{adaptive\textunderscore beta}}{{$\beta_{\textnormal{meta}}$, 
$\texttt{dist}_{\textnormal{meta}, v}$,
$\varepsilon$,
$\{\texttt{dist}_{\textnormal{meta}, s, r}\}^{s, r \in S, I}$,
$\mathbb{T}$, 
$K$, 
$lr_{\textnormal{base}}$}}
\If{$\varepsilon=\texttt{None}$}
    \State $\{ \theta_{\textnormal{base}, t} \}
    \leftarrow \texttt{base\textunderscore params}(\theta, \mathbb{T}, K, lr_{\textnormal{base}})$ %
    \State $\textnormal{dist\textunderscore list} = \{\}$ 
    \For {$\theta^{'}$ in $\{ \theta_{\textnormal{base}, t} \}$}
        \State $\{ \theta^{'}_{\textnormal{base}, t} \}
    \leftarrow \texttt{base\textunderscore params}(\theta^{'}, \mathbb{T}, K, lr_{\textnormal{base}})$
    \State $\textnormal{dist\textunderscore list} \leftarrow \frac{1}{T} \Sigma_{t}^{T} \texttt{dist}(\theta^{'}_{\textnormal{base}, t}, \theta_{\textnormal{base}, t})$
    \EndFor
    \State $\varepsilon = \max(\textnormal{dist\textunderscore list})$
\EndIf
\State $\texttt{dist}_{\textnormal{meta}} := \arg\max_{s, r} [ \texttt{dist}_{\textnormal{meta}, s, r} | \texttt{dist}_{\textnormal{base}, s, r} \leq \varepsilon ]$ %
\State $\beta_{\textnormal{meta}} = \max(\beta_{\textnormal{meta}}, \beta_{\textnormal{meta}} \times \frac{\texttt{dist}_{\textnormal{meta}}}{\texttt{dist}_{\textnormal{meta}}-\texttt{dist}_{\textnormal{meta}, v}} ) $
\State \textbf{return} $\beta_{\textnormal{meta}}$
\EndProcedure
\end{algorithmic}
\label{alg:beta}
\end{algorithm}

\section{Subnetwork Projection (SNP): Expanding projected subspace to support drift}

Given a model architecture, 
we can alter the training procedure to one of gradient-based meta learning and project the subnetworks onto a base learner's parameter space. 
In the standard implementation, we assume no memory $\mathcal{M}=\texttt{False}$. 
We cannot access subnetworks, but we can regulate the training of the meta parameters such that 
we can maximize the radius of the parameter subspace in which the meta parameter can drift, while returning the same base parameters within bounded error (Algorithms \ref{alg:train}-\ref{alg:expand}).
Referring to Algorithm \ref{alg:train}, per epoch,
after computing the support set loss w.r.t. computed base parameters, we compute a set of distance regularization terms. 
Our selected distance function $\texttt{dist}$ is cosine distance. 
We sample interpolated meta parameters at varying distances from the current epoch’s meta parameters, and compute the cumulative drift in meta parameters and base parameters. 
With these loss terms, we update the meta parameters. 
In an online setting (Algorithm \ref{alg:expand}), 
we perform distance regularization on the meta parameters (but not the base parameters, as $\mathcal{M}=\texttt{False}$).
Given knowledge of the subspace radius from the training procedure, while we measure the drift of the meta parameters, we are informed on when the base parameters error will increase (e.g. exceeding the radius). As such, we make use of an adaptive regularization coefficient procedure (Algorithm \ref{alg:beta}): when meta parameters are closer to the end of the supported radius, the distance regularization coefficient will increase accordingly.

\section{SNP++: Memory-based subnetwork access and manipulation}

\begin{algorithm}[ht]
\small
\caption{{\bf \texttt{train\textunderscore space}}}
\begin{algorithmic}[1]
\Procedure{\texttt{train\textunderscore space}}{$\mathbb{T} = \{ \mathcal{T}_t \}^{t \in T}$,
$K = 5$, 
$\texttt{epochs}=10,000$,
$lr_{\textnormal{base}}=0.0005$,
$lr_{\textnormal{meta}}=0.0005$,
$\beta_{\textnormal{meta}}=0.5$,
$\beta_{\textnormal{base}}=\{\beta_{\textnormal{base}, t}=0.5\}^{t \in T}$, 
$S=1,000$,
$I = [0.0001, 0.001, 0.01, 0.1]$, 
$\mathcal{M}=\texttt{False} \textnormal{ or } \{\}$}
\State $\theta \leftarrow \theta_{\textnormal{init}}$
\If{$\mathcal{M} \neq \texttt{False}$} \Comment{optional: store memory}
    \State $\mathcal{M} \leftarrow \{X^K_t, Y^K_t \}^{t \in T}$
\EndIf
\For{\texttt{epoch} in \texttt{epochs}}
    \State $\{ \theta_{\textnormal{base}, t} \}
    \leftarrow \texttt{base\textunderscore params}(\theta, \mathbb{T}, K, lr_{\textnormal{base}})$ %
    \For{$\mathcal{T}_t$ in $\mathbb{T}$} %
        \For {$X_t$, $Y_t$ in $\mathcal{T}_t$}
            \State $\mathcal{L}_{\mathcal{T}_t} =  \mathcal{L}(\theta - lr_{\textnormal{base}} \frac{\partial \mathcal{L}(\theta; X_t, Y_t)}{\partial \theta}; X_t, Y_t)$
        \EndFor
    \EndFor
    \For{$s$ in $S$} 
        \State $\theta_{\textnormal{rand}} \leftarrow \theta_{\textnormal{init}}$
        \For{$r$ in $I$}
            \State $\theta_{\textnormal{int}} = (1-r) \theta + r \theta_{\textnormal{rand}}$ %
            \State $\texttt{dist}_{\textnormal{meta}, s, r} = \texttt{dist}(\theta, \theta_{\textnormal{int}})$ %
            \State $\{ \theta^{\textnormal{int}}_{\textnormal{base}, t} \} \leftarrow \texttt{base\textunderscore params}(\theta_{\textnormal{int}}, \mathbb{T}, K, lr_{\textnormal{base}})$
            \State $\texttt{dist}_{\textnormal{base}, s, r} = \Sigma_t^T \texttt{dist}(\theta_{\textnormal{base}, t}, \theta^{\textnormal{int}}_{\textnormal{base}, t})$ 
        \EndFor
    \EndFor
    \State $\mathcal{L}_{\textnormal{meta}} = \Sigma_s^S \Sigma_r^I \texttt{dist}_{\textnormal{meta}, s, r}$
    \State 
    $\mathcal{L}_{\textnormal{base}} = \Sigma_s^S \Sigma_r^I \texttt{dist}_{\textnormal{base}, s, r}$
    \State $\theta := \theta 
    - lr_{\textnormal{meta}} \Sigma_t^T \frac{\partial \mathcal{L}_{\mathcal{T}_t}}{\partial \theta}
    - \beta_{\textnormal{meta}} \frac{\partial \mathcal{L}_{\textnormal{meta}}}{\partial \theta}
    - \beta_{\textnormal{base}} \frac{\partial \mathcal{L}_{\textnormal{base}}}{\partial \theta}$ %
\EndFor
\State \textbf{return} $\theta, \mathcal{M}$
\EndProcedure
\end{algorithmic}
\label{alg:train}
\end{algorithm}
\begin{algorithm}[ht]
\small
\caption{{\bf \texttt{expand\textunderscore space}}}
\begin{algorithmic}[1]
\Procedure{\texttt{expand\textunderscore space}}{$\theta$,
$\mathbb{V} = \{ \mathcal{V}_v \}^{v \in V}$,
$K = 5$,
$\texttt{epochs}=500$,
$lr_{\textnormal{base}}=0.0005$,
$lr_{\textnormal{meta}}=0.0005$
$\beta_{\textnormal{meta}}=0.5$,
$\beta_{\textnormal{base}}=\{\beta_{\textnormal{base}, v}=0.5\}^{v \in V}$, 
$\beta_{\textnormal{int}}=\{\beta_{\textnormal{int}, v}=1.0\}^{v \in V}$, 
$\mathcal{M}=\texttt{False} \textnormal{ or } \{X^K_t, Y^K_t \}^{t \in T}$,
$\varepsilon = 0.001 \textnormal{ or } \texttt{None}$}
\If{$\mathcal{M} \neq \texttt{False}$} \Comment{optional: access subnetworks}
    \For{$X^K_t, Y^K_t$ in $\mathcal{M}$}
        \State $\theta^{*}_{\textnormal{base}, t} = \theta - lr_{\textnormal{base}} \frac{\partial \mathcal{L}(\theta; X^K_t, Y^K_t)}{\partial \theta}$
    \EndFor
\EndIf
\For{$\mathcal{V}_v$ in $\mathbb{V}$}
    \For{\texttt{epoch} in \texttt{epochs}}
        \State $\{ \theta_{\textnormal{base}, v} \} \leftarrow \texttt{base\textunderscore params}(\theta, \mathbb{V}, K, lr_{\textnormal{base}})$ %
        \If{$\mathcal{M} \neq \texttt{False}$} %
            \If{$\beta_{\textnormal{int}, v} > 0$}
                \For{$\theta_{\textnormal{base}, v}$ in $\{\theta_{\textnormal{base}, v}\}$}
                    \State $g := \arg\min_{g \in T} \texttt{dist}(\theta_{\textnormal{base}, v}, \theta^{*}_{\textnormal{base}, g})$ %
                \EndFor
            \EndIf
        \EndIf
        \For{$X_v, Y_v$ in $\mathcal{V}_v$} %
            \State $\mathcal{L}_{\mathcal{V}_v} =  \mathcal{L}(\theta - lr_{\textnormal{base}} \frac{\partial \mathcal{L}(\theta; X_v, Y_v)}{\partial \theta}; X_v, Y_v)$
        \EndFor
        \State $\texttt{dist}_{\textnormal{meta}, v} = \texttt{dist}(\theta, \theta - lr_{\textnormal{meta}} \Sigma_v^V \frac{\partial \mathcal{L}_{\mathcal{V}_v}}{\partial \theta} )$ %
        \If{$\mathcal{M} \neq \texttt{False}$} %
            \For{$X^K_t, Y^K_t$ in $\mathcal{M}$}
                \If{$\beta_{\textnormal{base}, t} > 0$} %
        \State 
       $\texttt{dist}_{\textnormal{base}, t} = \texttt{dist}(\theta^{*}_{\textnormal{base}, t}, \theta - lr_{\textnormal{base}} \frac{\partial \mathcal{L}(\theta; X^K_t, Y^K_t)}{\partial \theta} )$
                \EndIf
            \EndFor
        \EndIf
        \If{$\mathcal{M} \neq \texttt{False}$} \Comment{optional: interp./remove}
            \If{$\beta_{\textnormal{int}, v} > 0$}
                \State $X^K_g, Y^K_g \leftarrow \mathcal{M}$
                \State $X^K_v, Y^K_v \leftarrow \mathcal{V}_v$
    \State 
       $\texttt{dist}_{\textnormal{int}} = \texttt{dist}(\theta - lr_{\textnormal{base}} \frac{\partial \mathcal{L}(\theta; X^K_v, Y^K_v)}{\partial \theta}, \theta - lr_{\textnormal{base}} \frac{\partial \mathcal{L}(\theta; X^K_g, Y^K_g)}{\partial \theta})$
            \EndIf
        \EndIf
    \EndFor
    \State $\beta_{\textnormal{meta}} \leftarrow \texttt{adaptive\textunderscore beta}(\beta_{\textnormal{meta}}, \texttt{dist}_{\textnormal{meta}, v}, \varepsilon)$
    \State $\mathcal{L}_{\textnormal{meta}} = \texttt{dist}_{\textnormal{meta}, v}$; $\mathcal{L}_{\textnormal{int}} = \texttt{dist}_{\textnormal{int}}$
    \State $\mathcal{L}_{\textnormal{base}} = \Sigma_t^T \texttt{dist}_{\textnormal{base}, t}$
    \State 
        $\theta := \theta 
    - lr_{\textnormal{meta}} \Sigma_v^V \frac{\partial \mathcal{L}_{\mathcal{V}_v}}{\partial \theta} - \beta_{\textnormal{meta}} \frac{\partial \mathcal{L}_{\textnormal{meta}}}{\partial \theta}
    - \beta_{\textnormal{base}} \frac{\partial \mathcal{L}_{\textnormal{base}}}{\partial \theta}
    - \beta_{\textnormal{int}, g} \frac{\partial \mathcal{L}_{\textnormal{int}}}{\partial \theta}$
    \If{$\mathcal{M} \neq \texttt{False}$} %
        \State $\mathcal{M} \leftarrow \{X^K_v, Y^K_v \}^{v \in V}$
    \EndIf
\EndFor
\State \textbf{return} $\theta, \mathcal{M}$
\EndProcedure
\end{algorithmic}
\label{alg:expand}
\end{algorithm}

To query a subnetwork, we need task-specific data, in-line with prior subnetwork literature.
Unlike replay-based methods, we do not store extensive replay buffers; instead, the memory buffer is one instance of a N-way-K-shot support set for computing base parameters. 
The use of this task-specific support introduces various interesting properties for manipulating the model.
As the support set varies, we can map each input distribution to a unique subnetwork. As such, we have a continuous space 
of subnetworks. 
In the standard case, we intend to add new subnetworks. First we use the previous training procedure to maximize the subspace radius. 
For each new task, we can fine-tune our meta parameters w.r.t. the new dataset, while using the memory buffer to track and regularize the drift of each individual base parameter. 
Unlike SNP, we regularize both the drift in the meta parameters as well as each base parameter.

Extending further from subnetwork addition, we also can evaluate subnetwork removal, combining (or interpolating between) subnetworks, and even switching subnetworks to alternate subnetwork modes. 
For subnetwork removal, we can choose not to freeze/regularize a specific task's subnetwork (e.g. setting its regularization coefficient to be less then 1.0 for partial removal, or even setting to 0.0 to ignore its regularization). 
It does not actively remove the subnetwork, but it also does not actively preserve it.
A use case is if a particular subnetwork causes interference, or if the capacity is needed for another task. In these cases, a new task’s base parameter can overwrite this base parameter.

For interpolating between subnetworks, other than adding a new subnetwork entirely, we can save capacity and allow one task's base parameters to be used in multiple tasks. 
We can first evaluate which existing base parameter is closest to the new task, and use this is the target base parameter. 
Then we can update the meta parameters such that, while the meta parameters drift and other non-target base parameter drift is minimized, 
the target base parameter is being updated towards the new task while performing well on its prior tasks. 

For mode switching, the parameter space has many functionally-diverse modes that we may wish to use and replace an existing subnetwork in-place. 
For example, we could replace a task's base parameter with an adversarially-robust parameter (e.g. sampled with \citet{https://doi.org/10.48550/arxiv.1412.6572, datta-2022-learn2weight}), or a backdoor-robust parameter (e.g sampled with \citet{https://doi.org/10.48550/arxiv.2102.13624, https://doi.org/10.48550/arxiv.2201.12211, https://doi.org/10.48550/arxiv.2203.03692}), or domain-robust parameters, etc. 
Rather than using the task's original training set to locate this mode, an alternative approach would actively sample the parameter space, and update the meta parameters and regularize the drift of the replaced mode's base parameter such that the new base paramater is computed with respect to the specific task. 
While it is possible to actively sample the base parameters iteratively to identify the ideal base parameter mode \citep{https://doi.org/10.48550/arxiv.2205.09891}, 
it poses a risk that the target mode may cause the meta parameter to drift beyond the subspace radius.
Thus, for our evaluation of the identification of a sharpness-aware mode (for low-loss basins using SAM \citep{foret2021sharpnessaware}), 
we actively sample meta parameters graduating from within the radius to outside of the radius, and for each sampled meta parameter we compute the base parameter, and evaluate whether it satisfies the mode's evaluation condition (e.g. flat basin).

\section{Experiments}

\subsection{Methodology}
\label{exp}

\textbf{Model. }
We evaluate with CLIP \citep{radford2021learning}, the standard vision-language model, 
specifically the pre-trained initialization of the ResNet-50 variant. 
For training/fine-tuning on a new task, 
we retain the 
Adam optimizer, 
decoupled weight decay regularization,
temperature clipping,
and a batch size of 32.
From the pre-trained initialization, we train CLIP on MSCOCO for 50 epochs
until both loss convergence and verification that zero/few-shot performance across the datasets is weakened. 
We fine-tune for 10 epochs and also validate loss convergence. 

\textbf{Adaptation baselines. }
Fine-tuning (Single Task Learning) is the default baseline in
the online/continual learning setting, 
where the model sequentially learns each incoming task without any forgetting mitigation.
Joint Training (Multi Task Learning) is a baseline where the model can train on all future tasks at once. We do not include the base task (MSCOCO), and evaluate when there are at least 2 tasks (Task 3). 

We evaluate against 5 baseline adaptation methods, 
including
3 general-purpose continual learning strategies, 
and 2 large-model-specific adaptation strategies (that have also been evaluated on CLIP).
Elastic Weight Consolidation (EWC) \citep{EWC} uses weight regularization to retain weights from previous tasks. The regularization strength for weight penalty $\lambda$ is $1,000$.
Gradient Projection Memory (GPM) \citep{saha2021gradient} learns new tasks by performing gradient steps in the orthogonal direction to the gradient subspaces that are important to the past tasks.
We use a 0.01 learning rate.
BatchEnsemble \citep{Wen2020BatchEnsemble}
uses a base network (slow weights) and stores separate parameters (fast weights) to compute the parameters per ensemble, thus $N$ ensembles do not require $N$ sets of parameters. Each ensemble member is responsible for one task.
We retain -0.5 random sign init for fast weights and 0.5 fast weights learning rate multiplier.

\begin{table*}[t]
  \centering
  \caption{
  Moving from pre-trained initialization, to Task 1-3, we present the (Zero-shot Top-5 / Few-shot Top-1) accuracy across each task and baseline method.
We also measure backward transfer (BWT) \citep{lopez2017gradient}, which is the influence that learning a new task has on the performance on a previous task.
Positive backward transfer occurs when learning a new task increases the performance on a preceding task. Negative backward transfer occurs when learning about a new task decreases the performance on a preceding task.
Rather than computing a general BWT score, 
for those tasks that are greater than or equal to its compared value, we compute the average positive BWT. 
For those tasks that are less than its compared value, we compute the average negative BWT. 
This helps us measure positive transfer as well as drawdown.
For Task 3, we evaluate each method against their method's task 2 performance; 
otherwise, each task's method is evaluated against the previous task's fine-tuning performance.
  }
  \label{tab:baselines}
    \resizebox{\textwidth}{!}{
    \begin{tabular}{|l|c|cccccc|ccc|}
    \hline \hline
          & Task 1: MSCOCO & \multicolumn{9}{c|}{Task 2: ImageNet}                                  \\     \hline
& Fine-tuning 
& Fine-tuning 
& EWC
& BatchEnsemble
& GPM
& CLIP-Adapter
& PAINT
& SNP
& SNP++ (Add.)
& SNP++ (Inter.)
\\ \hline
MSCOCO (N=91) & 93.7 / 91.2 & 51.2 / 54.8 & 78.3 / 84.7 & 87.4 / 86.6 & 77.5 / 86.8 & 80.5 / 81.6 & 86.0 / 89.9 & 88.2 / 89.3 & 90.8 / 86.0 & 86.3 / 92.4  \\
    ImageNet (N=1000) & 8.6 / 10.9 & 83.2 / 31.3 & 71.4 / 28.6 & 74.6 / 29.0 & 72.7 / 28.6 & 75.9 / 27.6 & 81.0 / 31.3 & 78.6 / 36.4 & 80.9 / 35.1 & 76.9 / 37.7  \\
    CIFAR100 (N=100) & 11.3 / 10.8 & 15.8 / 9.2 & 11.5 / 10.2 & 11.4 / 10.4 & 11.4 / 10.1 & 11.7 / 10.4 & 15.0 / 10.7 & 14.7 / 11.3 & 15.1 / 10.9 & 14.4 / 11.7  \\
    STL10 (N=10) & 95.3 / 55.2 & 97.2 / 61.7 & 95.4 / 54.9 & 91.3 / 56.2 & 88.0 / 57.6 & 96.3 / 56.6 & 96.1 / 55.9 & 94.5 / 52.4 & 97.3 / 50.5 & 92.5 / 54.2  \\
    Caltech101 (N=102) & 8.1 / 43.4 & 8.2 / 50.5 & 8.3 / 46.8 & 8.5 / 44.9 & 7.7 / 46.2 & 9.1 / 45.0 & 8.2 / 45.1 & 9.4 / 47.9 & 9.7 / 46.1 & 9.2 / 49.6  \\
    StanfordCars (N=196) & 2.7 / 7.2 & 3.1 / 7.3 & 3.4 / 7.3 & 2.2 / 7.3 & 3.6 / 7.2 & 3.5 / 7.1 & 3.9 / 7.3 & 3.7 / 7.6 & 3.8 / 7.3 & 3.6 / 7.9 \\
    Flowers102 (N=102) & 7.0 / 55.1 & 6.7 / 60.8 & 6.5 / 56.3 & 6.8 / 59.0 & 6.7 / 54.8 & 7.1 / 55.2 & 6.9 / 59.4 & 7.4 / 63.7 & 7.6 / 61.4 & 7.2 / 65.9  \\
    GTSRB (N=43) & 17.2 / 16.6 & 15.6 / 17.2 & 16.7 / 16.7 & 16.2 / 17.1 & 16.7 / 17.2 & 16.9 / 17.2 & 15.1 / 17.2 & 16.1 / 16.9 & 16.6 / 16.3 & 15.8 / 17.5  \\
    Food101 (N=101) & 16.4 / 12.3 & 13.7 / 8.5 & 14.1 / 10.1 & 15.9 / 11.4 & 15.5 / 9.8 & 15.0 / 11.1 & 16.3 / 10.5 & 15.8 / 11.2 & 16.3 / 10.8 & 15.5 / 11.6  \\
    EuroSAT (N=10) & 54.3 / 52.4 & 51.4 / 60.1 & 53.2 / 54.8 & 51.8 / 57.3 & 53.5 / 56.9 & 48.1 / 52.8 & 51.1 / 60.5 & 55.7 / 59.1 & 57.4 / 56.9 & 54.5 / 61.1  \\
    FGVCAircraft (N=100) & 5.4 / 7.8 & 5.6 / 10.6 & 5.5 / 8.4 & 5.6 / 9.3 & 5.3 / 7.4 & 5.5 / 7.7 & 6.0 / 10.4 & 6.3 / 8.8 & 6.5 / 8.5 & 6.2 / 9.1 \\ \hline
    Avg   & 29.1 / 33.0 & 32.0 / 33.8 & 33.1 / 34.4 & 33.8 / 35.3 & 32.6 / 34.8 & 33.6 / 33.9 & 35.1 / 36.2 & 35.5 / 36.8 & 36.5 / 35.4 & 34.7 / 38.1  \\
    Pos BWT &   0.0 / 0.0 &	13.6 / 6.4 &	10.7 / 3.6 &	16.7 / 3.9 &	21.7 / 5.6 &	10.1 / 3.5 &	13.1 / 4.8 &	11.2 / 5.9 &	10.7 / 5.5	& 10.7 / 6.4
    \\
    Neg BWT  & 0.0 / 0.0 &	-10.0 / -13.9 &	-4.0 / -2.4 &	-2.1 / -2.0 &	-3.3 / -1.4 &	-5.3 / -2.3 &	-2.6 / -1.1 &	-2.0 / -1.9 &	-1.2 / -2.9 &	-3.1 / -0.8
    \\
    \hline 
    \end{tabular}%
    }
  \resizebox{\textwidth}{!}{
    \begin{tabular}{|l|ccccccc|ccc|}
    \hline
          & \multicolumn{10}{c|}{Task 3: Caltech101} \\
\hline       
& Fine-tuning
& Joint Training
& EWC
& BatchEnsemble
& GPM
& CLIP-Adapter
& PAINT
& SNP
& SNP++ (Add.)
& SNP++ (Inter.) \\ \hline
MSCOCO (N=91) & 29.1 / 37.8 & 37.3 / 39.4 & 68.6 / 69.4 & 84.2 / 86.7 & 71.7 / 73.6 & 76.7 / 79.2 & 71.1 / 71.9 & 78.2 / 81.3 & 79.0 / 80.1 & 71.5 / 83.1 \\
    ImageNet (N=1000) & 15.7 / 12.9 & 80.9 / 28.6 & 63.7 / 24.9 & 71.1 / 24.6 & 67.1 / 21.8 & 70.3 / 25.4 & 77.0 / 25.0 & 76.5 / 27.3 & 77.3 / 26.9 & 69.9 / 27.9 \\
    CIFAR100 (N=100) & 6.9 / 12.7 & 12.1 / 12.0 & 10.3 / 9.8 & 10.1 / 8.8 & 7.8 / 12.2 & 10.1 / 9.2 & 14.2 / 8.5 & 13.8 / 10.2 & 13.9 / 10.0 & 12.6 / 10.4 \\
    STL10 (N=10) & 91.5 / 45.3 & 95.2 / 62.9 & 73.1 / 38.7 & 83.6 / 47.3 & 88.2 / 47.8 & 91.4 / 53.9 & 91.5 / 44.7 & 90.6 / 46.4 & 91.6 / 45.7 & 82.8 / 47.4 \\
    Caltech101 (N=102) & 96.1 / 93.7 & 88.2 / 89.7 & 85.0 / 92.2 & 86.5 / 85.0 & 88.8 / 84.2 & 87.1 / 82.2 & 82.4 / 70.2 & 89.1 / 91.2 & 90.1 / 89.8 & 81.4 / 93.3 \\
    StanfordCars (N=196) & 2.9 / 4.6 & 3.2 / 6.3 & 3.1 / 3.6 & 2.2 / 6.1 & 3.1 / 5.1 & 2.9 / 5.3 & 3.6 / 5.8 & 3.6 / 6.0 & 3.6 / 5.9 & 3.3 / 6.1 \\
    Flowers102 (N=102) & 4.7 / 47.8 & 5.8 / 61.6 & 5.9 / 43.6 & 6.6 / 48.0 & 5.1 / 49.2 & 5.6 / 55.5 & 6.6 / 47.5 & 5.9 / 50.1 & 6.0 / 49.3 & 5.4 / 51.2 \\
    GTSRB (N=43) & 10.5 / 9.1 & 14.7 / 16.9 & 11.6 / 10.6 & 15.5 / 10.6 & 11.7 / 10.7 & 12.4 / 10.6 & 14.4 / 13.8 & 13.9 / 14.2 & 14.1 / 14.0 & 12.7 / 14.5 \\
    Food101 (N=101) & 8.2 / 6.3 & 11.3 / 10.0 & 6.7 / 7.0 & 13.6 / 9.5 & 9.7 / 7.0 & 14.6 / 11.1 & 15.7 / 8.4 & 15.5 / 9.8 & 15.7 / 9.7 & 14.2 / 10.0 \\
    EuroSAT (N=10) & 48.8 / 46.4 & 53.9 / 62.6 & 41.9 / 56.5 & 51.5 / 52.1 & 49.8 / 48.5 & 49.5 / 56.5 & 50.8 / 48.4 & 54.4 / 55.7 & 55.0 / 54.9 & 49.7 / 57.0 \\
    FGVCAircraft (N=100) & 5.4 / 6.8 & 5.4 / 9.3 & 6.0 / 6.9 & 5.3 / 7.7 & 5.4 / 7.0 & 5.6 / 7.7 & 5.8 / 8.3 & 6.1 / 8.1 & 6.2 / 8.0 & 5.6 / 8.3 \\
    \hline
    Avg   & 29.1 / 29.4 & 37.1 / 36.3 & 34.2 / 33.0 & 39.1 / 35.1 & 37.1 / 33.4 & 38.7 / 36.1 & 39.4 / 32.0 & 40.7 / 36.4 & 41.1 / 35.8 & 37.2 / 37.2 
    \\
    Pos BWT &   87.9 / 23.3 &	27.6 / 8.0 &	38.6 / 23.6 &	39.1 / 20.1 &	27.1 / 20.0 &	26.5 / 8.2 &	74.2 / 25.1 &	79.7 / 43.3 &	80.4 / 43.7 &	72.2 / 43.7
    \\
    Neg BWT  & -12.0 / -10.6 &	-3.3 / -4.2 &	-7.3 / -7.0 &	-2.2 / -4.7 &	-4.0 / -6.2 &	-2.9 / -2.8 &	-2.7 / -7.1 &	-2.3 / -4.8 &	-3.0 / -3.9 &	-4.5 / -5.3
    \\
    \hline \hline
    \end{tabular}%
    }
\end{table*}%

CLIP-Adapter \citep{gao2021clip}
fine-tunes with feature adapters,
specifically an additional bottleneck layer to learn new features and perform residual-style feature blending with the original pre-trained features.
In-line with the implementation for CLIP, 
we fine tune the visual adapter.
PAINT \citep{ilharcopatching} is another vision-language model adaptation technique. 
It is a patching method that interpolates between parameters before fine-tuning and parameters after fine-tuning on a task to be patched.
We implement sequential patching, where we iteratively repeat the patching procedure on each new task, 
and pick the mixing coefficient that optimizes average accuracy on the held-out validation sets from the supported and patching tasks.
All baselines are trained with the CLIP model.

\textbf{Zero-shot. }
Unlike traditional visual systems trained on a fixed set of discrete labels, \citet{radford2021learning} shared a new paradigm in learning to align images with raw texts in an open-vocabulary setting. 
For zero-shot classification, all class labels are first converted to sentences using their prompt templates, and the model computes their text embeddings. 
To classify an image, the model computes the image embedding,
and then an image-text similarity score (cosine similarity) is computed between the image embedding and each class' text embeddings to find strongest alignment. 
The calculated similarity of these embeddings per class, scaled by a temperature parameter, is then normalized into a probability distribution via softmax.
The class having the highest similarity with the image is the predicted one.
We perform zero-shot evaluation with the altered CLIP-Adapter model, and perform zero-shot evaluation with the task-indexed BatchEnsemble model. 
For all other methods, albeit differences in the training procedure, the model architecture and parameters are available for directly applying this zero-shot evaluation procedure.

\textbf{Few-shot. }
For a given task, an N-way-K-shot support set (N classes, K samples per class) is provided to assist the method to make inferences.
Then evaluation is performed on the query set. 
For meta learners, 
the meta learner computes the base parameters with respect to the support set.
Specific to gradient-based meta learners, including MAML and SNP(++), we compute the gradient of the support set with respect to model parameters, update the model parameters, then evaluate on the query set. 

For standard CLIP and the other methods, 
we use nearest-neighbor classification.
We first compute the mean image features per class in the support set, 
then measure the (cosine) distance between them and the image features for a given query set image.
Based on the nearest class mean, the closest class' mean image features is the predicted class.

\begin{table*}
\centering
\caption{
Variations in hyperparameters and subnetwork manipulation strategies with SNP(++). 
}
\includegraphics[width=\linewidth]{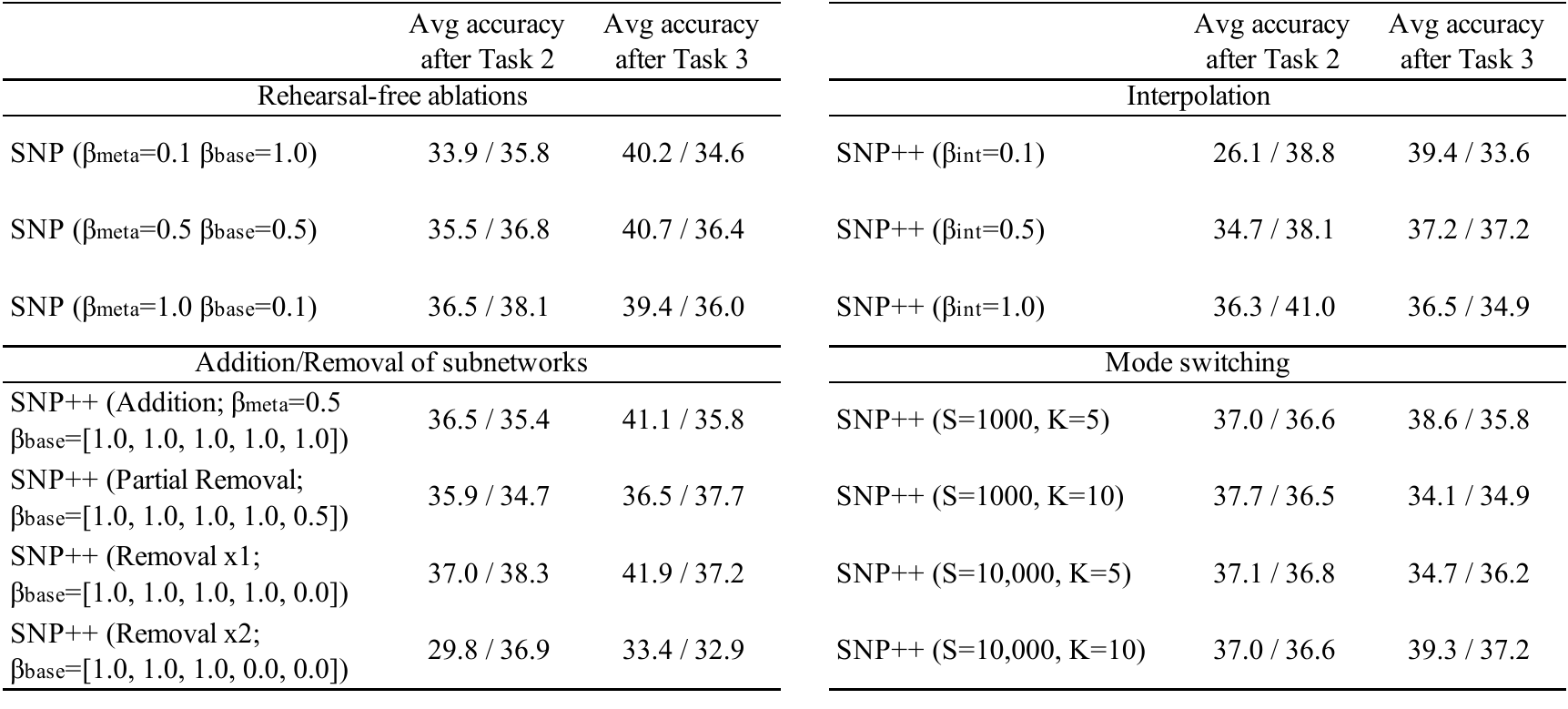}
\label{fig:ablations}
\end{table*}

\subsection{Maintaining zero/few-shot capabilities}

We tabulate our proposed method in comparison to baselines in Table \ref{tab:baselines}, and 
proposed method in comparison to different configurations in Table \ref{fig:ablations}.

Transferability between tasks plays an important role.
From task 1 to 3, we find that positive backward transfer exists across all baselines, and that some datasets have improved zero/few-shot performance with task shift. 
Furthermore, we find that the gradual removal of subnetworks with SNP++ may worsen performance. The removal of subnetworks is motivated by alleviating interference between task-specific representations between two tasks; in this case, it appears that the attempt at removal overwrites the transferable representations. 
Further sub-procedures that identify the optimal subnetworks to remove based on a transfer-interference trade-off can improve the utility of subnetwork removal, especially in a setting with many tasks. 

Adaptation techniques specialized for large models (CLIP in particular) outperform general-purpose continual learning methods. 
For large models, 
regularization-based methods that do not require task indexing or separate context vectors can perform competitively to non-regularization-based methods. 

Our proposed adaptation method, SNP and SNP++, outperforms existing baselines. 
It can consistently retain low negative backward transfer, fulfilling the objective of minimizing loss of zero/few-shot performance with task shift. 
It performs comparably in maximizing positive backward transfer. 
In terms of balancing between positive and negative backward transfer, SNP and SNP++ strikes the optimal balance, attaining the highest average accuracy. 

We find that our proposed method works better with the memory buffer that regularizes the base parameter drift.
Though we are not storing trajectories or large replay buffers (only storing one support set instance), 
pure-regularization SNP can also perform in a stable fashion.
Different hyperparameters of $\beta_{\textnormal{base}}$ and $\beta_{\textnormal{meta}}$ tend to retain similar performances, and no major loss in accuracy is observed. 
We do note that setting the $\beta_{\textnormal{meta}}$ too low can worsen performance, particularly in comparison to non-SNP baselines. This may occur where the drift of the meta parameters is under-regulated, and regulating base parameter drift is insufficient and acts as a second-order regularizer of the meta parameter drift.

We find that combining subnetworks and interpolating between them underperforms SNP and additive SNP++. 
Subnetwork addition/removal would manipulate the number of base parameters, 
but would result in first-order interpolation between the unchanged meta parameters and new meta parameters with the modified subnetwork set. 
Thus, interpolating between subnetworks results in second-order interpolation, 
and the error with respect to each task accumulates when the meta parameters interpolate. 
Given a very large number of tasks and lower model capacity, second-order interpolation offers an efficient subnetwork manipulation.

Subnetwork projection improves the maintainance of zero/few-shot capabilities in an online setting. Moreover, subnetwork projection supports the manipulation of task-specific representations for many more applications.

\section{Conclusion}
\label{6}

By projecting a model's subnetworks onto the same equidimensional parameter space as the (meta) parameters, 
we can edit the representations encoded in the network, including adding, removing, combining, and switching subnetworks. 
We apply this paradigm to achieve superior online/continual learning performance in retaining seen and zero/few-shot accuracy on prior and subsequent tasks. 
Not only does our method scale to large models, 
it also lays the foundation for further network editing applications, 
such as subnetwork removal for privacy (e.g. machine unlearning), or subnetwork addition for distributional robustness (e.g. adding distributionally-varied samples for fairness or adversarial robustness).

\clearpage
\bibliography{main}

\begin{thebibliography}{49}
\providecommand{\natexlab}[1]{#1}
\providecommand{\url}[1]{\texttt{#1}}
\expandafter\ifx\csname urlstyle\endcsname\relax
  \providecommand{\doi}[1]{doi: #1}\else
  \providecommand{\doi}{doi: \begingroup \urlstyle{rm}\Url}\fi

\bibitem[Andreas et~al.(2016{\natexlab{a}})Andreas, Rohrbach, Darrell, and
  Klein]{Andreas_2016_CVPR}
Jacob Andreas, Marcus Rohrbach, Trevor Darrell, and Dan Klein.
\newblock Neural module networks.
\newblock In \emph{Proceedings of the IEEE Conference on Computer Vision and
  Pattern Recognition (CVPR)}, June 2016{\natexlab{a}}.

\bibitem[Andreas et~al.(2016{\natexlab{b}})Andreas, Rohrbach, Darrell, and
  Klein]{https://doi.org/10.48550/arxiv.1601.01705}
Jacob Andreas, Marcus Rohrbach, Trevor Darrell, and Dan Klein.
\newblock Learning to compose neural networks for question answering,
  2016{\natexlab{b}}.
\newblock URL \url{https://arxiv.org/abs/1601.01705}.

\bibitem[Bansal et~al.(2021)Bansal, Nakkiran, and Barak]{bansal2021revisiting}
Yamini Bansal, Preetum Nakkiran, and Boaz Barak.
\newblock Revisiting model stitching to compare neural representations.
\newblock In A.~Beygelzimer, Y.~Dauphin, P.~Liang, and J.~Wortman Vaughan
  (eds.), \emph{Advances in Neural Information Processing Systems}, 2021.
\newblock URL \url{https://openreview.net/forum?id=ak06J5jNR4}.

\bibitem[Bossard et~al.(2014)Bossard, Guillaumin, and Van~Gool]{bossard14}
Lukas Bossard, Matthieu Guillaumin, and Luc Van~Gool.
\newblock Food-101 -- mining discriminative components with random forests.
\newblock In \emph{European Conference on Computer Vision}, 2014.

\bibitem[Brown et~al.(2020)Brown, Mann, Ryder, Subbiah, Kaplan, Dhariwal,
  Neelakantan, Shyam, Sastry, Askell, Agarwal, Herbert-Voss, Krueger, Henighan,
  Child, Ramesh, Ziegler, Wu, Winter, Hesse, Chen, Sigler, Litwin, Gray, Chess,
  Clark, Berner, McCandlish, Radford, Sutskever, and
  Amodei]{https://doi.org/10.48550/arxiv.2005.14165}
Tom~B. Brown, Benjamin Mann, Nick Ryder, Melanie Subbiah, Jared Kaplan,
  Prafulla Dhariwal, Arvind Neelakantan, Pranav Shyam, Girish Sastry, Amanda
  Askell, Sandhini Agarwal, Ariel Herbert-Voss, Gretchen Krueger, Tom Henighan,
  Rewon Child, Aditya Ramesh, Daniel~M. Ziegler, Jeffrey Wu, Clemens Winter,
  Christopher Hesse, Mark Chen, Eric Sigler, Mateusz Litwin, Scott Gray,
  Benjamin Chess, Jack Clark, Christopher Berner, Sam McCandlish, Alec Radford,
  Ilya Sutskever, and Dario Amodei.
\newblock Language models are few-shot learners, 2020.
\newblock URL \url{https://arxiv.org/abs/2005.14165}.

\bibitem[Chen et~al.(2022)Chen, GE, Tong, Wang, Song, Wang, and
  Luo]{chen2022adaptformer}
Shoufa Chen, Chongjian GE, Zhan Tong, Jiangliu Wang, Yibing Song, Jue Wang, and
  Ping Luo.
\newblock Adaptformer: Adapting vision transformers for scalable visual
  recognition.
\newblock In Alice~H. Oh, Alekh Agarwal, Danielle Belgrave, and Kyunghyun Cho
  (eds.), \emph{Advances in Neural Information Processing Systems}, 2022.
\newblock URL \url{https://openreview.net/forum?id=ATiz_CDA66}.

\bibitem[Chiaro et~al.(2020)Chiaro, Twardowski, Bagdanov, and van~de
  Weijer]{10.5555/3495724.3497128}
Riccardo~Del Chiaro, Bartlomiej Twardowski, Andrew~D. Bagdanov, and Joost
  van~de Weijer.
\newblock Ratt: Recurrent attention to transient tasks for continual image
  captioning.
\newblock In \emph{Proceedings of the 34th International Conference on Neural
  Information Processing Systems}, NIPS'20, Red Hook, NY, USA, 2020. Curran
  Associates Inc.
\newblock ISBN 9781713829546.

\bibitem[Coates et~al.(2011)Coates, Ng, and Lee]{coates2011stl10}
Adam Coates, Andrew Ng, and Honglak Lee.
\newblock {An Analysis of Single Layer Networks in Unsupervised Feature
  Learning}.
\newblock In \emph{AISTATS}, 2011.
\newblock
  \url{https://cs.stanford.edu/~acoates/papers/coatesleeng_aistats_2011.pdf}.

\bibitem[Csiszárik et~al.(2021)Csiszárik, Kőrösi-Szabó, Matszangosz, Papp,
  and Varga]{https://doi.org/10.48550/arxiv.2110.14633}
Adrián Csiszárik, Péter Kőrösi-Szabó, Akos~K. Matszangosz, Gergely Papp,
  and Dániel Varga.
\newblock Similarity and matching of neural network representations, 2021.
\newblock URL \url{https://arxiv.org/abs/2110.14633}.

\bibitem[Datta(2022)]{datta-2022-learn2weight}
Siddhartha Datta.
\newblock {L}earn2{W}eight: Parameter adaptation against similar-domain
  adversarial attacks.
\newblock In \emph{Proceedings of the 29th International Conference on
  Computational Linguistics}, pp.\  4832--4843, Gyeongju, Republic of Korea,
  October 2022. International Committee on Computational Linguistics.
\newblock URL \url{https://aclanthology.org/2022.coling-1.427}.

\bibitem[Datta \& Shadbolt(2022{\natexlab{a}})Datta and
  Shadbolt]{https://doi.org/10.48550/arxiv.2201.12211}
Siddhartha Datta and Nigel Shadbolt.
\newblock Backdoors stuck at the frontdoor: Multi-agent backdoor attacks that
  backfire, 2022{\natexlab{a}}.
\newblock URL \url{https://arxiv.org/abs/2201.12211}.

\bibitem[Datta \& Shadbolt(2022{\natexlab{b}})Datta and
  Shadbolt]{https://doi.org/10.48550/arxiv.2203.03692}
Siddhartha Datta and Nigel Shadbolt.
\newblock Low-loss subspace compression for clean gains against multi-agent
  backdoor attacks, 2022{\natexlab{b}}.
\newblock URL \url{https://arxiv.org/abs/2203.03692}.

\bibitem[Datta \& Shadbolt(2022{\natexlab{c}})Datta and
  Shadbolt]{https://doi.org/10.48550/arxiv.2205.09891}
Siddhartha Datta and Nigel Shadbolt.
\newblock Interpolating compressed parameter subspaces, 2022{\natexlab{c}}.
\newblock URL \url{https://arxiv.org/abs/2205.09891}.

\bibitem[Datta \& Shadbolt(2022{\natexlab{d}})Datta and
  Shadbolt]{https://doi.org/10.48550/arxiv.2209.14996}
Siddhartha Datta and Nigel Shadbolt.
\newblock Multiple modes for continual learning, 2022{\natexlab{d}}.
\newblock URL \url{https://arxiv.org/abs/2209.14996}.

\bibitem[Deng et~al.(2009)Deng, Dong, Socher, Li, Li, and
  Fei-Fei]{deng2009imagenet}
Jia Deng, Wei Dong, Richard Socher, Li-Jia Li, Kai Li, and Li~Fei-Fei.
\newblock Imagenet: A large-scale hierarchical image database.
\newblock In \emph{CVPR}, pp.\  248--255. Ieee, 2009.

\bibitem[Fei-Fei et~al.(2004)Fei-Fei, Fergus, and Perona]{FeiFei2004LearningGV}
Li~Fei-Fei, Rob Fergus, and Pietro Perona.
\newblock Learning generative visual models from few training examples: An
  incremental bayesian approach tested on 101 object categories.
\newblock \emph{Computer Vision and Pattern Recognition Workshop}, 2004.

\bibitem[Finn et~al.(2017{\natexlab{a}})Finn, Abbeel, and
  Levine]{finn2017modelagnostic}
Chelsea Finn, Pieter Abbeel, and Sergey Levine.
\newblock Model-agnostic meta-learning for fast adaptation of deep networks,
  2017{\natexlab{a}}.

\bibitem[Finn et~al.(2017{\natexlab{b}})Finn, Abbeel, and
  Levine]{pmlr-v70-finn17a}
Chelsea Finn, Pieter Abbeel, and Sergey Levine.
\newblock Model-agnostic meta-learning for fast adaptation of deep networks.
\newblock In Doina Precup and Yee~Whye Teh (eds.), \emph{Proceedings of the
  34th International Conference on Machine Learning}, volume~70 of
  \emph{Proceedings of Machine Learning Research}, pp.\  1126--1135. PMLR,
  06--11 Aug 2017{\natexlab{b}}.
\newblock URL \url{http://proceedings.mlr.press/v70/finn17a.html}.

\bibitem[Foret et~al.(2021)Foret, Kleiner, Mobahi, and
  Neyshabur]{foret2021sharpnessaware}
Pierre Foret, Ariel Kleiner, Hossein Mobahi, and Behnam Neyshabur.
\newblock Sharpness-aware minimization for efficiently improving
  generalization.
\newblock In \emph{International Conference on Learning Representations}, 2021.
\newblock URL \url{https://openreview.net/forum?id=6Tm1mposlrM}.

\bibitem[Gao et~al.(2021)Gao, Geng, Zhang, Ma, Fang, Zhang, Li, and
  Qiao]{gao2021clip}
Peng Gao, Shijie Geng, Renrui Zhang, Teli Ma, Rongyao Fang, Yongfeng Zhang,
  Hongsheng Li, and Yu~Qiao.
\newblock Clip-adapter: Better vision-language models with feature adapters.
\newblock \emph{arXiv preprint arXiv:2110.04544}, 2021.

\bibitem[Geiping et~al.(2021)Geiping, Fowl, Somepalli, Goldblum, Moeller, and
  Goldstein]{https://doi.org/10.48550/arxiv.2102.13624}
Jonas Geiping, Liam Fowl, Gowthami Somepalli, Micah Goldblum, Michael Moeller,
  and Tom Goldstein.
\newblock What doesn't kill you makes you robust(er): How to adversarially
  train against data poisoning, 2021.
\newblock URL \url{https://arxiv.org/abs/2102.13624}.

\bibitem[Goodfellow et~al.(2014)Goodfellow, Shlens, and
  Szegedy]{https://doi.org/10.48550/arxiv.1412.6572}
Ian~J. Goodfellow, Jonathon Shlens, and Christian Szegedy.
\newblock Explaining and harnessing adversarial examples, 2014.
\newblock URL \url{https://arxiv.org/abs/1412.6572}.

\bibitem[Ha et~al.(2016)Ha, Dai, and Le]{ha2016hypernetworks}
David Ha, Andrew Dai, and Quoc~V Le.
\newblock Hypernetworks.
\newblock \emph{arXiv preprint arXiv:1609.09106}, 2016.

\bibitem[Helber et~al.(2017)Helber, Bischke, Dengel, and
  Borth]{helber2017eurosat}
Patrick Helber, Benjamin Bischke, Andreas Dengel, and Damian Borth.
\newblock Eurosat: A novel dataset and deep learning benchmark for land use and
  land cover classification, 2017.

\bibitem[Hoffmann et~al.(2022)Hoffmann, Borgeaud, Mensch, Buchatskaya, Cai,
  Rutherford, Casas, Hendricks, Welbl, Clark, et~al.]{hoffmann2022training}
Jordan Hoffmann, Sebastian Borgeaud, Arthur Mensch, Elena Buchatskaya, Trevor
  Cai, Eliza Rutherford, Diego de~Las Casas, Lisa~Anne Hendricks, Johannes
  Welbl, Aidan Clark, et~al.
\newblock Training compute-optimal large language models.
\newblock \emph{arXiv preprint arXiv:2203.15556}, 2022.

\bibitem[Hospedales et~al.(2021)Hospedales, Antoniou, Micaelli, and
  Storkey]{hospedales2021meta}
Timothy Hospedales, Antreas Antoniou, Paul Micaelli, and Amos Storkey.
\newblock Meta-learning in neural networks: A survey.
\newblock \emph{IEEE transactions on pattern analysis and machine
  intelligence}, 44\penalty0 (9):\penalty0 5149--5169, 2021.

\bibitem[Ilharco et~al.(2022)Ilharco, Wortsman, Gadre, Song, Hajishirzi,
  Kornblith, Farhadi, and Schmidt]{ilharcopatching}
Gabriel Ilharco, Mitchell Wortsman, Samir~Yitzhak Gadre, Shuran Song, Hannaneh
  Hajishirzi, Simon Kornblith, Ali Farhadi, and Ludwig Schmidt.
\newblock Patching open-vocabulary models by interpolating weights.
\newblock In \emph{Advances in Neural Information Processing Systems}, 2022.

\bibitem[Kang et~al.(2022)Kang, Mina, Madjid, Yoon, Hasegawa-Johnson, Hwang,
  and Yoo]{pmlr-v162-kang22b}
Haeyong Kang, Rusty John~Lloyd Mina, Sultan Rizky~Hikmawan Madjid, Jaehong
  Yoon, Mark Hasegawa-Johnson, Sung~Ju Hwang, and Chang~D. Yoo.
\newblock Forget-free continual learning with winning subnetworks.
\newblock In Kamalika Chaudhuri, Stefanie Jegelka, Le~Song, Csaba Szepesvari,
  Gang Niu, and Sivan Sabato (eds.), \emph{Proceedings of the 39th
  International Conference on Machine Learning}, volume 162 of
  \emph{Proceedings of Machine Learning Research}, pp.\  10734--10750. PMLR,
  17--23 Jul 2022.
\newblock URL \url{https://proceedings.mlr.press/v162/kang22b.html}.

\bibitem[Kirkpatrick et~al.(2017)Kirkpatrick, Pascanu, Rabinowitz, Veness, and
  et. al.]{EWC}
James~N Kirkpatrick, Razvan Pascanu, Neil~C. Rabinowitz, Joel Veness, and et.
  al.
\newblock Overcoming catastrophic forgetting in neural networks.
\newblock \emph{Proceedings of the National Academy of Sciences of the United
  States of America}, 114 13:\penalty0 3521--3526, 2017.

\bibitem[Krause et~al.(2013)Krause, Stark, Deng, and
  Fei-Fei]{KrauseStarkDengFei-Fei_3DRR2013}
Jonathan Krause, Michael Stark, Jia Deng, and Li~Fei-Fei.
\newblock 3d object representations for fine-grained categorization.
\newblock In \emph{4th International IEEE Workshop on 3D Representation and
  Recognition (3dRR-13)}, Sydney, Australia, 2013.

\bibitem[{Krizhevsky}(2009)]{krizhevsky2009learning}
Alex {Krizhevsky}.
\newblock Learning multiple layers of features from tiny images.
\newblock 2009.

\bibitem[Lange et~al.(2019)Lange, Aljundi, Masana, Parisot, Jia, Leonardis,
  Slabaugh, and Tuytelaars]{Lange2019ContinualLA}
Matthias Lange, Rahaf Aljundi, Marc Masana, Sarah Parisot, Xu~Jia, Ale
  Leonardis, Gregory~G. Slabaugh, and Tinne Tuytelaars.
\newblock Continual learning: A comparative study on how to defy forgetting in
  classification tasks.
\newblock \emph{ArXiv}, abs/1909.08383, 2019.

\bibitem[Lin et~al.(2014)Lin, Maire, Belongie, Bourdev, Girshick, Hays, Perona,
  Ramanan, Zitnick, and Dollár]{https://doi.org/10.48550/arxiv.1405.0312}
Tsung-Yi Lin, Michael Maire, Serge Belongie, Lubomir Bourdev, Ross Girshick,
  James Hays, Pietro Perona, Deva Ramanan, C.~Lawrence Zitnick, and Piotr
  Dollár.
\newblock Microsoft coco: Common objects in context, 2014.
\newblock URL \url{https://arxiv.org/abs/1405.0312}.

\bibitem[Lopez-Paz \& Ranzato(2017)Lopez-Paz and Ranzato]{lopez2017gradient}
David Lopez-Paz and Marc'Aurelio Ranzato.
\newblock Gradient episodic memory for continual learning.
\newblock In \emph{Advances in Neural Information Processing Systems}, pp.\
  6467--6476, 2017.

\bibitem[Maji et~al.(2013)Maji, Kannala, Rahtu, Blaschko, and
  Vedaldi]{maji13fine-grained}
S.~Maji, J.~Kannala, E.~Rahtu, M.~Blaschko, and A.~Vedaldi.
\newblock Fine-grained visual classification of aircraft.
\newblock Technical report, 2013.

\bibitem[Matena \& Raffel(2021)Matena and
  Raffel]{https://doi.org/10.48550/arxiv.2111.09832}
Michael Matena and Colin Raffel.
\newblock Merging models with fisher-weighted averaging, 2021.
\newblock URL \url{https://arxiv.org/abs/2111.09832}.

\bibitem[Nilsback \& Zisserman(2008)Nilsback and Zisserman]{Nilsback08}
M-E. Nilsback and A.~Zisserman.
\newblock Automated flower classification over a large number of classes.
\newblock In \emph{Proceedings of the Indian Conference on Computer Vision,
  Graphics and Image Processing}, Dec 2008.

\bibitem[Radford et~al.(2021)Radford, Kim, Hallacy, Ramesh, Goh, Agarwal,
  Sastry, Askell, Mishkin, Clark, Krueger, and Sutskever]{radford2021learning}
Alec Radford, Jong~Wook Kim, Chris Hallacy, Aditya Ramesh, Gabriel Goh,
  Sandhini Agarwal, Girish Sastry, Amanda Askell, Pamela Mishkin, Jack Clark,
  Gretchen Krueger, and Ilya Sutskever.
\newblock Learning transferable visual models from natural language
  supervision, 2021.

\bibitem[Saha et~al.(2021)Saha, Garg, and Roy]{saha2021gradient}
Gobinda Saha, Isha Garg, and Kaushik Roy.
\newblock Gradient projection memory for continual learning.
\newblock In \emph{International Conference on Learning Representations}, 2021.
\newblock URL \url{https://openreview.net/forum?id=3AOj0RCNC2}.

\bibitem[Sanh et~al.(2022)Sanh, Webson, Raffel, Bach, Sutawika, Alyafeai,
  Chaffin, Stiegler, Raja, Dey, Bari, Xu, Thakker, Sharma, Szczechla, Kim,
  Chhablani, Nayak, Datta, Chang, Jiang, Wang, Manica, Shen, Yong, Pandey,
  Bawden, Wang, Neeraj, Rozen, Sharma, Santilli, Fevry, Fries, Teehan, Scao,
  Biderman, Gao, Wolf, and Rush]{sanh2022multitask}
Victor Sanh, Albert Webson, Colin Raffel, Stephen Bach, Lintang Sutawika, Zaid
  Alyafeai, Antoine Chaffin, Arnaud Stiegler, Arun Raja, Manan Dey, M~Saiful
  Bari, Canwen Xu, Urmish Thakker, Shanya~Sharma Sharma, Eliza Szczechla,
  Taewoon Kim, Gunjan Chhablani, Nihal Nayak, Debajyoti Datta, Jonathan Chang,
  Mike Tian-Jian Jiang, Han Wang, Matteo Manica, Sheng Shen, Zheng~Xin Yong,
  Harshit Pandey, Rachel Bawden, Thomas Wang, Trishala Neeraj, Jos Rozen,
  Abheesht Sharma, Andrea Santilli, Thibault Fevry, Jason~Alan Fries, Ryan
  Teehan, Teven~Le Scao, Stella Biderman, Leo Gao, Thomas Wolf, and Alexander~M
  Rush.
\newblock Multitask prompted training enables zero-shot task generalization.
\newblock In \emph{International Conference on Learning Representations}, 2022.
\newblock URL \url{https://openreview.net/forum?id=9Vrb9D0WI4}.

\bibitem[Singh \& Jaggi(2019)Singh and
  Jaggi]{https://doi.org/10.48550/arxiv.1910.05653}
Sidak~Pal Singh and Martin Jaggi.
\newblock Model fusion via optimal transport, 2019.
\newblock URL \url{https://arxiv.org/abs/1910.05653}.

\bibitem[Stallkamp et~al.(2012)Stallkamp, Schlipsing, Salmen, and
  Igel]{Stallkamp2012}
J.~Stallkamp, M.~Schlipsing, J.~Salmen, and C.~Igel.
\newblock Man vs. computer: Benchmarking machine learning algorithms for
  traffic sign recognition.
\newblock \emph{Neural Networks}, \penalty0 (0):\penalty0 --, 2012.
\newblock ISSN 0893-6080.
\newblock \doi{10.1016/j.neunet.2012.02.016}.
\newblock URL
  \url{http://www.sciencedirect.com/science/article/pii/S0893608012000457}.

\bibitem[Wei et~al.(2022{\natexlab{a}})Wei, Bosma, Zhao, Guu, Yu, Lester, Du,
  Dai, and Le]{wei2022finetuned}
Jason Wei, Maarten Bosma, Vincent Zhao, Kelvin Guu, Adams~Wei Yu, Brian Lester,
  Nan Du, Andrew~M. Dai, and Quoc~V Le.
\newblock Finetuned language models are zero-shot learners.
\newblock In \emph{International Conference on Learning Representations},
  2022{\natexlab{a}}.
\newblock URL \url{https://openreview.net/forum?id=gEZrGCozdqR}.

\bibitem[Wei et~al.(2022{\natexlab{b}})Wei, Tay, Bommasani, Raffel, Zoph,
  Borgeaud, Yogatama, Bosma, Zhou, Metzler, Chi, Hashimoto, Vinyals, Liang,
  Dean, and Fedus]{wei2022emergent}
Jason Wei, Yi~Tay, Rishi Bommasani, Colin Raffel, Barret Zoph, Sebastian
  Borgeaud, Dani Yogatama, Maarten Bosma, Denny Zhou, Donald Metzler, Ed~H.
  Chi, Tatsunori Hashimoto, Oriol Vinyals, Percy Liang, Jeff Dean, and William
  Fedus.
\newblock Emergent abilities of large language models.
\newblock \emph{Transactions on Machine Learning Research}, 2022{\natexlab{b}}.
\newblock URL \url{https://openreview.net/forum?id=yzkSU5zdwD}.
\newblock Survey Certification.

\bibitem[Wen et~al.(2020)Wen, Tran, and Ba]{Wen2020BatchEnsemble}
Yeming Wen, Dustin Tran, and Jimmy Ba.
\newblock Batchensemble: an alternative approach to efficient ensemble and
  lifelong learning.
\newblock In \emph{International Conference on Learning Representations}, 2020.
\newblock URL \url{https://openreview.net/forum?id=Sklf1yrYDr}.

\bibitem[Wortsman et~al.(2022)Wortsman, Ilharco, Gadre, Roelofs, Gontijo-Lopes,
  Morcos, Namkoong, Farhadi, Carmon, Kornblith, and
  Schmidt]{https://doi.org/10.48550/arxiv.2203.05482}
Mitchell Wortsman, Gabriel Ilharco, Samir~Yitzhak Gadre, Rebecca Roelofs,
  Raphael Gontijo-Lopes, Ari~S. Morcos, Hongseok Namkoong, Ali Farhadi, Yair
  Carmon, Simon Kornblith, and Ludwig Schmidt.
\newblock Model soups: averaging weights of multiple fine-tuned models improves
  accuracy without increasing inference time, 2022.
\newblock URL \url{https://arxiv.org/abs/2203.05482}.

\bibitem[Yoon et~al.(2020)Yoon, Kim, Yang, and Hwang]{Yoon2020Scalable}
Jaehong Yoon, Saehoon Kim, Eunho Yang, and Sung~Ju Hwang.
\newblock Scalable and order-robust continual learning with additive parameter
  decomposition.
\newblock In \emph{International Conference on Learning Representations}, 2020.
\newblock URL \url{https://openreview.net/forum?id=r1gdj2EKPB}.

\bibitem[Zenke et~al.(2017)Zenke, Poole, and Ganguli]{10.5555/3305890.3306093}
Friedemann Zenke, Ben Poole, and Surya Ganguli.
\newblock Continual learning through synaptic intelligence.
\newblock In \emph{Proceedings of the 34th International Conference on Machine
  Learning - Volume 70}, ICML'17, pp.\  3987–3995. JMLR.org, 2017.

\bibitem[Zhou et~al.(2022)Zhou, Yang, Loy, and Liu]{Zhou_2022}
Kaiyang Zhou, Jingkang Yang, Chen~Change Loy, and Ziwei Liu.
\newblock Learning to prompt for vision-language models.
\newblock \emph{International Journal of Computer Vision}, 130\penalty0
  (9):\penalty0 2337--2348, jul 2022.
\newblock \doi{10.1007/s11263-022-01653-1}.
\newblock URL \url{https://doi.org/10.1007%2Fs11263-022-01653-1}.

\end{thebibliography}
\bibliographystyle{iclr2023_conference}

\end{document}